\def\year{2020}\relax
\documentclass[letterpaper]{article} 
\usepackage{aaai20}  
\usepackage{times}  
\usepackage{helvet} 
\usepackage{courier}  
\usepackage[hyphens]{url}  
\usepackage{graphicx} 
\usepackage{graphicx}  
\usepackage{amssymb}
\usepackage{amsmath}
\usepackage{footnote}
\usepackage{multirow}
\usepackage{booktabs}
\usepackage{threeparttable}
\usepackage{enumerate}
\usepackage{indentfirst}
\usepackage{array}
\usepackage{rotating}
\usepackage{ragged2e}
\usepackage{wasysym}

\def\ie{{\em i.e.}}

\def\etal{{\em et al.}}

\title{Ultrafast Video Attention Prediction with Coupled Knowledge Distillation}
\author{Kui Fu\textsuperscript{\rm 1}, Peipei Shi\textsuperscript{\rm 2}, Yafei Song\textsuperscript{\rm 3}, Shiming Ge\textsuperscript{\rm 4}, Xiangju Lu\textsuperscript{\rm 2} and Jia Li\textsuperscript{\rm 1}\thanks{Correspondence should be addressed to Jia Li (E-mail: jiali@buaa.edu.cn) and Xiangju Lu (E-mail: luxiangju@qiyi.com).
}\\  
\textsuperscript{\rm 1}State Key Laboratory of Virtual Reality Technology and Systems, SCSE, Beihang University\\
\textsuperscript{\rm 2}iQIYI, Inc\\
\textsuperscript{\rm 3}National Engineering Laboratory for Video Technology, School of EE\&CS, Peking University \\
\textsuperscript{\rm 4}Institute of Information Engineering, Chinese Academy of Sciences
}
 
\begin{document}

\maketitle

\begin{abstract}
Large convolutional neural network models have recently demonstrated impressive performance on video attention prediction. Conventionally, these models are with intensive computation and large memory. To address these issues, we design an extremely light-weight network with ultrafast speed, named UVA-Net. The network is constructed based on depth-wise convolutions and takes low-resolution images as input. However, this straight-forward acceleration method will decrease performance dramatically. To this end, we propose a coupled knowledge distillation strategy to augment and train the network effectively. With this strategy, the model can further automatically discover and emphasize implicit useful cues contained in the data. Both spatial and temporal knowledge learned by the high-resolution complex teacher networks also can be distilled and transferred into the proposed low-resolution light-weight spatiotemporal network. Experimental results show that the performance of our model is comparable to 11 state-of-the-art models in video attention prediction, while it costs only 0.68 MB memory footprint, runs about 10,106 FPS on GPU and 404 FPS on CPU, which is 206 times faster than previous models.
\end{abstract}

\section{Introduction}
\label{sec:intro}

Recent developments of portable/wearable devices have heightened the need for video attention prediction. Benefiting from comprehensive rules \cite{Li2015Finding}, large-scale training datasets \cite{deng2009imagenet} and deep learning algorithms \cite{Pan2016Shallow}, it becomes feasible to construct more and more complex models to improve the performance steadily in recent years. It is generally expected that such models could be used to facilitate subsequent tasks such as event understanding \cite{shu2015joint} and drone navigation \cite{zhang2010novel}.
\begin{figure}[t]
\begin{center}
   \includegraphics[width=1.0\columnwidth]{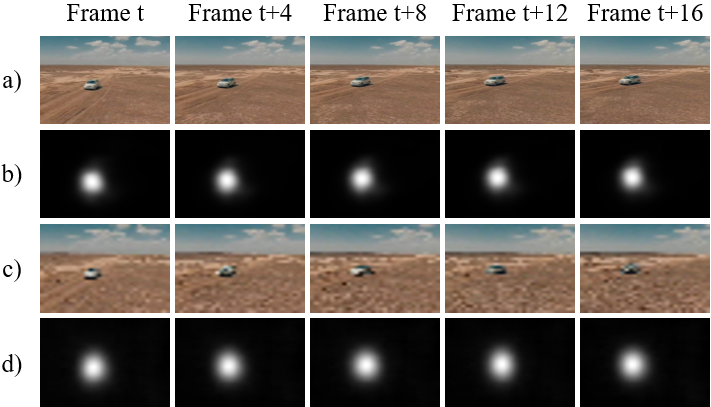}
\end{center}
   \caption{Comparison between the high-resolution and low-resolution videos. a) and b) are videos and their attention maps with high-resolution, while  c) and d) are with low-resolution.  Salient objects in low-resolution videos are missing much spatial information but still localizable.}
\label{fig:motivation}
\end{figure}

However, due to the limited computational ability and memory space of such portable/wearable devices, there are two main issues in video attention prediction: 1)~How to reduce the computational cost and memory space to enable the real-time applications on such devices? 2)~How to extract powerful spatiotemporal features from videos to avoid remarkable loss of attention prediction accuracy? To address these two issues, it is necessary to explore a feasible solution that converts existing complex and sparse spatial and temporal attention models into a simple and compact spatiotemporal network.

Over the past years, deep learning has become the dominant approach in attention prediction due to its impressive capability of handling large-scale learning problems \cite{wang2018deep}. Researchers tend to design more complex networks and collect more data for better performance. However, it has been proved that most of the complex networks are sparse and have a lot of redundancies. This fact facilitates the study of network compression, which aims to decrease computational and memory space cost. The video attention prediction can be greatly accelerated, but often at the expense of existing a prediction accuracy attenuation.

With the analyses in mind, we first reduce the resolution of the input image to $64 \times 64$ which decreases the computational and memory space cost by about one order of magnitude. As demonstrated in Fig. \ref{fig:motivation}, we observe that the salient objects in low-resolution videos are missing much spatial information but still localizable, which implies that it is still capable to recover the missing details. We then construct the network based on the depth-wise convolutions, which can further decrease the computational cost by about one order of magnitude and leads to an  Ultrafast Video Attention Prediction Network (UVA-Net). However, UVA-Net will suffer from a dramatic performance decrease with a straight-forward training strategy. To this end, we propose a coupled knowledge distillation strategy, which can augment the model to discover and emphasize useful cues from the data and extract spatiotemporal features simultaneously.

The contributions of this work are summarized as follows: 1) We design an extremely lightweight spatiotemporal network with an ultrafast speed, which is 206 times faster than previous methods. 2) We propose a coupled knowledge distillation strategy to enable the network to effectively fuse the spatial and temporal features simultaneously and avoid remarkable loss of attention prediction accuracy. 3) Comprehensive experiments are conducted and illustrate that our model can achieve an ultrafast speed with a comparable attention prediction accuracy to the state-of-the-art models.

\section{Related Work}
\label{sec:related}
In this section, we give a brief review of recent works from two perspectives: visual attention models as well as knowledge distillation and transfer.

\subsection{Visual Attention Models}

The heuristic models \cite{li2014saliency} can be roughly categorized into bottom-up approaches and top-down approaches. General models in the former category are stimulus-driven and compete to pop out visual signals. For example, Fang \etal~\shortcite{fang2014video} proposed a video attention model with a fusion strategy that according to the compactness and the temporal motion contrast. Later, Fang \etal~\shortcite{fang2017visual} proposed another model that based on the uncertainty weighting. The models in the latter category are task-driven and usually integrating high-level factors.

In order to overcome the deficiency of the heuristic model fusion strategy, researchers have proposed plenty of non-deep learning models \cite{li2010probabilistic}. Vig \etal~\shortcite{vig2012intrinsic} proposed a simple bottom-up attention model with supervised learning techniques fine-tune the free parameters for dynamic scenarios. Fang \etal~\shortcite{Fang2017Learning} proposed an optimization framework with pairwise binary terms to pop out salient targets and suppressing distractors.

For further performance improvements, some deep learning models are proposed, which are emphasizing the importance of automatic hierarchical feature extraction and end-to-end learning \cite{li2016deepsaliency}. For example, K\"{u}mmerer \etal~\shortcite{kummerer2016deepgaze} directly used the features from VGG-19 network \cite{simonyan2014very} for attention inference without additional fine-tuning. Pan \etal~\shortcite{Pan2016Shallow} proposed two networks for fixation prediction, with deep and shallow structures, respectively.

These modes can achieve impressive performance but usually have a high computational cost. How to obtain a speed-accuracy trade-off in attention prediction is still a key issue for scientific researchers.
\begin{figure*}[t]
\begin{center}
   \includegraphics[width=1.0\linewidth]{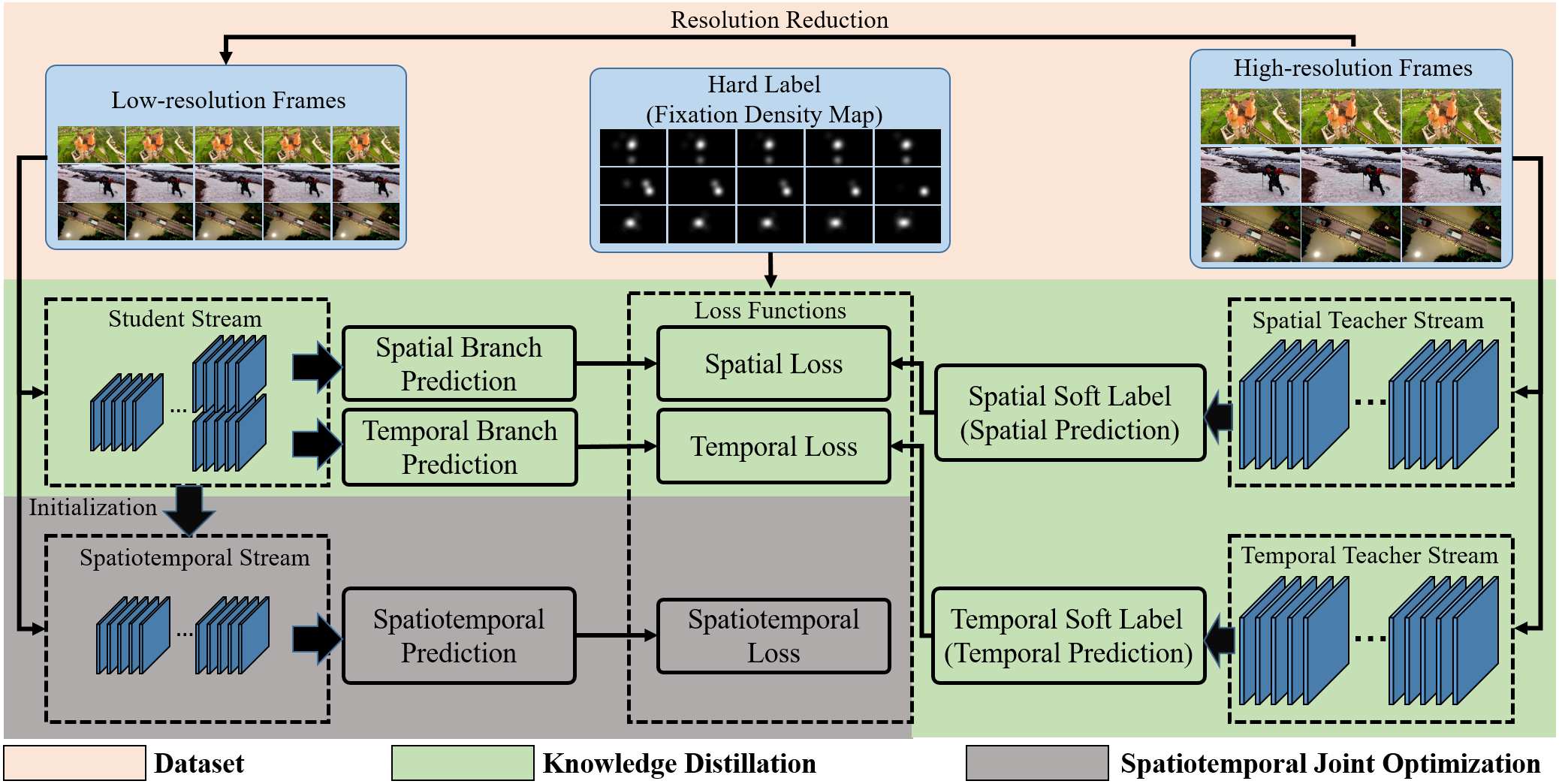}
\end{center}
   \caption{Overview. Our coupled knowledge distillation approach consists of four streams: a spatial teacher stream, a temporal teacher stream, a student stream and a spatiotemporal stream. It is trained in two steps: knowledge distillation and spatiotemporal joint optimization.}
\label{fig:framework}
\end{figure*}

\subsection{Knowledge Distillation and Transfer}
Knowledge distillation is a class of techniques originally proposed by Hinton \etal~\shortcite{hinton2015distilling}, which aims at transferring knowledge in complex teacher model to simple student model and improving the performance of student model at test time. In \cite{hinton2015distilling}, Hinton \etal~adopt the soft labels generated by teacher model as the supervision signal in addition to the regular labeled training data during the training phase. Studies have shown that such extra supervision signal can be in various reasonable forms, such as classification probabilities \cite{hinton2015distilling}, feature representation \cite{romero2014fitnets}, or inter-layer flow (the inner product of feature maps) \cite{yim2017gift}. It has observed a new wave of development exploiting knowledge transfer technologies to distill the knowledge in easy-to-train complex models into hard-to-train simple models.

There is no doubt that these knowledge distillation approaches are successful in high-resolution scenarios, but their effectiveness in low-resolution dynamic scenarios is questionable since they face the dual challenges of the limited network capacity and the loss of stimulus signal. With such questions, this paper demonstrates an ultrafast attention prediction network for videos.

\section{Coupled Knowledge Distillation}
\label{sec:method}
In this section, we present a coupled knowledge distillation approach for video attention prediction.

\subsection{Overview}
We start with an overview of our coupled knowledge distillation approach before going into details below. We learn a strong yet efficient attention predictor by transferring the knowledge of a spatial teacher network and a temporal teacher network to a simple and compact one.
The overview is as shown in Fig. \ref{fig:framework}.

The overall training process consists of two steps: 1) \textbf{Knowledge distillation.} Distilling the spatial and temporal knowledge inherited in the spatial teacher network and the temporal teacher network to a student network. 2) \textbf{Spatiotemporal joint optimization.} Transferring the knowledge learned by the student network to a spatiotemporal network and then fine-tune it.

\subsection{Residual Block with Channel-wise Attention}
\label{sec:attention}
\begin{figure}[t]
\begin{center}
   \includegraphics[width=1.0\columnwidth]{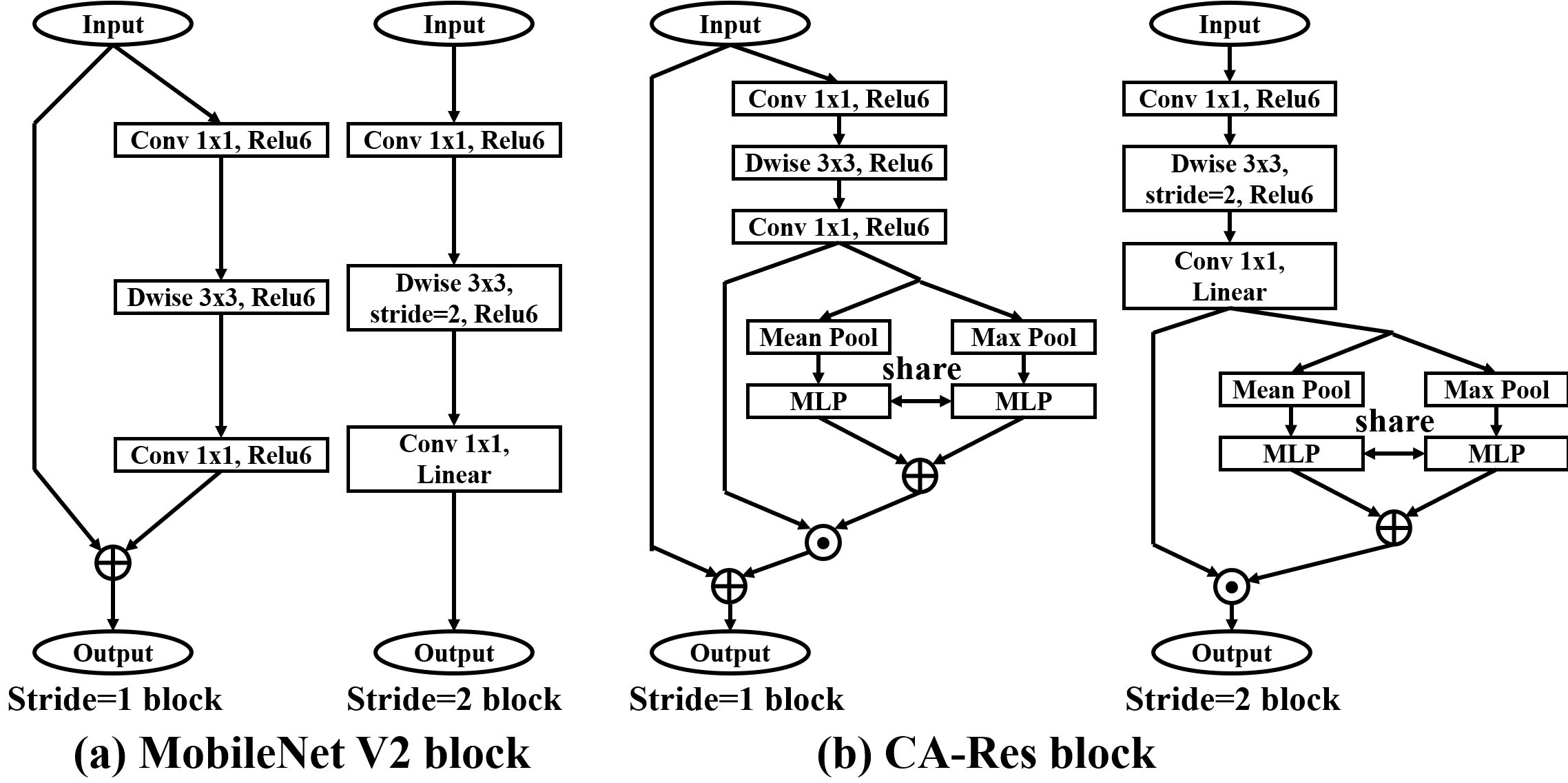}
\end{center}
   \caption{Detailed convolutional blocks. (a) MobileNet V2 block. (b) CA-Res block.}
\label{fig:conv_block}
\end{figure}

Inspired by MobileNetV2 \cite{sandler2018mobilenetv2}, we construct our networks with depth-wise convolutions with inverted residual structure. The basic convolutional blocks of MobileNet V2 are illustrated in Fig. \ref{fig:conv_block} (a). Such blocks can greatly improve the compactness of the networks but hard to maintain the same accuracy. Aiming to meet the requirements of practical applications on resource-limited devices, we introduce channel-wise attention mechanism in MobileNet V2 blocks and propose a novel light-weight block, named as CA-Res block. Detailed architecture of CA-Res blocks are as shown in Fig. \ref{fig:conv_block} (b).

Let $\textbf{R}_{in}$ and $\textbf{R}_{out}$ be the input and output intermediate representation, respectively. The $\textbf{R}_{out}$ can be formulated as
\begin{equation}
\textbf{R}_{out}=\textbf{R}_{in} + f^{inv}(\textbf{R}_{in})\odot f^{ca}(f^{inv}(\textbf{R}_{in})),
\end{equation}
where $f^{inv}$ denotes the inverted residual structure of MobileNet V2 block, $f^{ca}$ refers to a channel-wise attention function, $\odot$ is element-wise multiplication. We denote global average pooling and global max pooling as $P^{avg}$ and $P^{max}$, respectively. In general, $P^{avg}$ performs well in preserving global characteristics, while $P^{max}$ has the potential to remain the texture features \cite{woo2018cbam}. It has been proved that exploiting both of them can greatly improve the representation power of networks than using each independently. Given an intermediate feature map $\textbf{F}$, we can model $f^{ca}$ via
\begin{equation}
f^{ca}(\textbf{F})=\sigma(f^{MLP}(P^{avg}(\textbf{F}))+ f^{MLP}(P^{max}(\textbf{F}))),
\end{equation}
where $f^{MLP}$ refers to a multi-layer perceptron (MLP) which functioned as a generator of attention vectors, and $\sigma$ is sigmoid function. Both of $P^{avg}$ and $P^{max}$ are used to squeeze the spatial dimension of $\textbf{F}$.

\subsection{Knowledge Distillation}
Given the CA-Res block, we now describe the details to construct our knowledge distillation network. Suppose that the dataset is given as $\mathcal{D}=\{I_t,Y_t\}$, $t=1,2,3,\ldots,n$, where $I_t$ is an input image and $Y_t$ is a ground-truth map corresponding to $I_t$. Since this work aims to learn a spatiotemporal student model, the student architecture should be able to simulate the spatial/temporal features extracted by the teachers. Thus, our student network is a two-branch network that takes low-resolution successive frame pair $(I_t, I_{t+1})$ as input, and outputs both spatially and temporally predicted attentions. Note that in teacher networks, the temporal information is often extracted with the assistance of optical flows, as it provides effective ways to smoothly accumulate features along time. However, optical flows are expensive to compute, and hard to be applied in a lightweight model.

To effectively mimic temporal features with low cost, we propose to adopt simple operations on immediate features, which have more representation power, to mimic the temporal dynamics. Specifically, given the frame pair $(I_t, I_{t+1})$, the network first processes them through a low-level module $f^s$, which contains one stand convolutional layer and three CA-Res blocks in each branch, and shares the weights in the two paths. In this manner, the spatial intermediate representation $\mathcal{F}_t^{spa}$ and temporal intermediate representation $\mathcal{F}_t^{tem}$ can be computed via
\begin{eqnarray}
\mathcal{F}_t^{spa}&=&f^{s}(I_t),  \label{eq:f_1}\\
\mathcal{F}_t^{tem}&=&\mathrm{cat}\left(f^s(I_t), f^s(I_t) - f^s(I_{t+1})\right),  \label{eq:f_2}
\end{eqnarray}
where $\mathrm{cat}(\cdot)$ refers to a concatenating operation. We feed the $\mathcal{F}_t^{spa}$ into a spatial path, which consists of five blocks, to further extract high-level spatial intermediate representation $\mathcal{F}_t^{spa'}$. Similarly, the $\mathcal{F}_t^{tem}$ is fed into a temporal path to extract high-level temporal intermediate representation $\mathcal{F}_t^{tem'}$. Note that the spatial path and the temporal path share the same topology structure. Each path is followed with two deconvolutional layers to restore the $\mathcal{F}_t^{spa'}$ and $\mathcal{F}_t^{tem'}$ to fine feature map with the original resolution of input frames. The overall architecture of the student network is as shown in Fig. \ref{fig:network} (a).

We denote the spatial teacher, temporal teacher and student networks as $\mathbb{T}_{spa}$, $\mathbb{T}_{tem}$ and $\mathbb{S}$, respectively. Then, the $\mathbb{S}$ can be trained by optimizing spatial loss $\mathcal{L}_{spa}$ and temporal loss $\mathcal{L}_{tem}$
\begin{align}
\begin{split}
\mathcal{L}_{spa}=&(1-\mu)\cdot\mathcal{L}_{hard}(\mathbb{S}_{spa}(I_t,I_{t+1}),Y_t) \\
    &+\mu\cdot \mathcal{L}_{soft}(\mathbb{S}_{spa}(I_t,I_{t+1}),\mathbb{T}_{spa}(I_t)), \label{eq:l_s}
\end{split} \\
\begin{split}
\mathcal{L}_{tem}=&(1-\mu)\cdot\mathcal{L}_{hard}(\mathbb{S}_{tem}(I_t,I_{t+1}),Y_t) \\
    &+\mu\cdot \mathcal{L}_{soft}(\mathbb{S}_{tem}(I_t,I_{t+1}),\mathbb{T}_{tem}(I_t,I_{t+1})), \label{eq:l_t}
\end{split}
\end{align}
where $\mathcal{L}_{hard}$ and $\mathcal{L}_{soft}$ denote the hard loss and soft loss, respectively. The hard loss is a $\ell_2$ loss associated with the predicted density map and the ground truth density map, while the soft loss is a $\ell_2$ loss associated with the predicted density map and the teacher prediction following the practice in knowledge distillation \cite{hinton2015distilling}. The parameter $\mu$ balances $\mathcal{L}_{hard}$ and $\mathcal{L}_{soft}$ which we empirically set to $0.5$. $\mathbb{S}_{spa}$ and $\mathbb{S}_{tem}$ refer to the spatial and temporal branch of $\mathbb{S}$, respectively.

\subsection{Spatiotemporal Joint Optimization}
\begin{figure}[t]
\begin{center}
   \includegraphics[width=1.0\columnwidth]{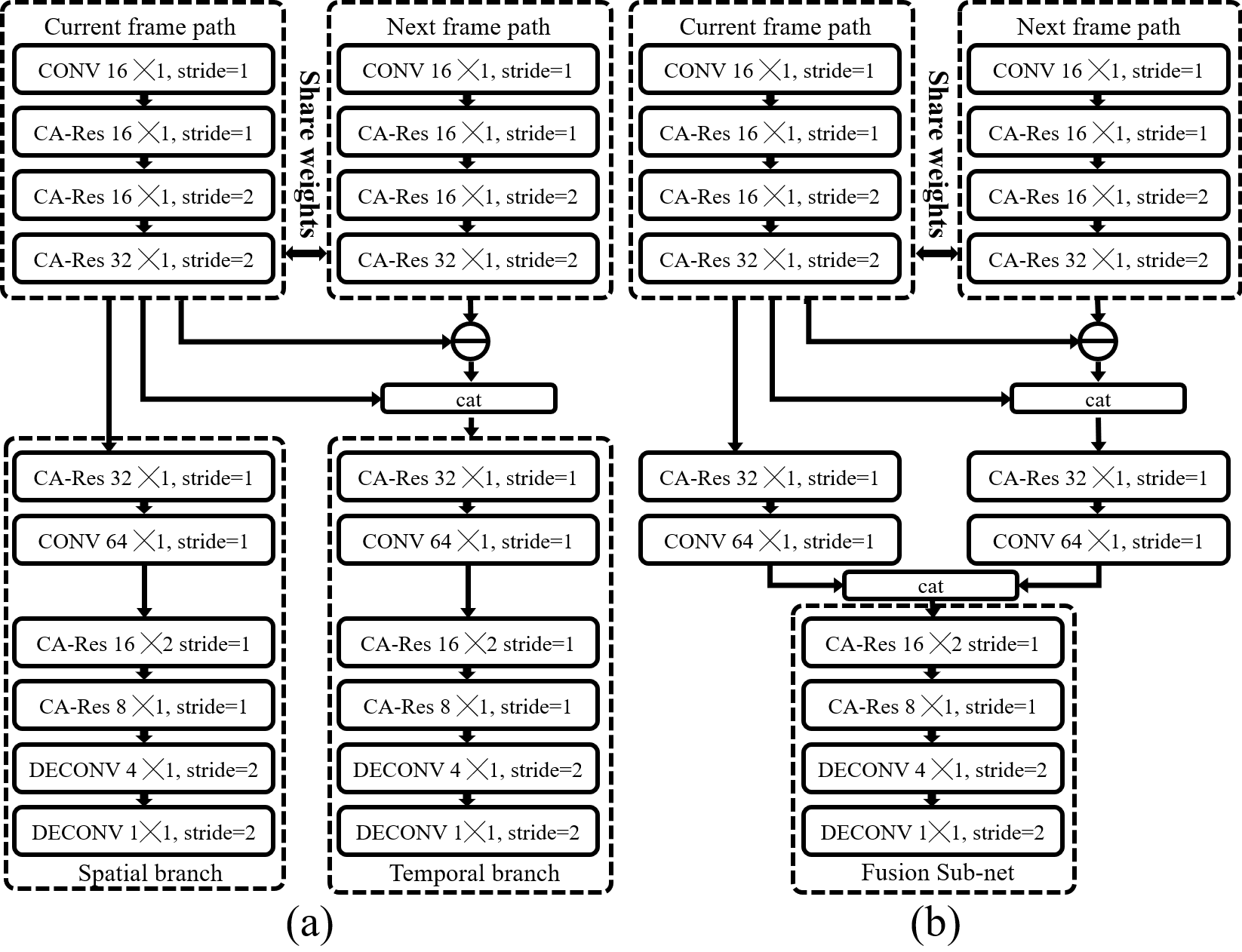}
\end{center}
   \caption{Network architecture. (a) Student network $\mathbb{S}$. (b) Spatiotemporal network $\mathbb{S}_{sp}$.}
\label{fig:network}
\end{figure}

Although the student model $\mathbb{S}$ can distill spatial and temporal knowledge inherited in teacher networks to itself separately, it is still challenging to fuse the spatial and temporal features together to provide more powerful representations. To address this issue, we construct a spatiotemporal network, which is denoted as $\mathbb{S}_{sp}$.  The first six blocks of $\mathbb{S}_{sp}$ share the same structure as with those of $\mathbb{S}$ to transferring the knowledge from the $\mathbb{S}$. In this manner, $\mathbb{S}_{sp}$ can generate both spatial intermediate representation $\mathcal{F}_t^{spa''}$ and temporal intermediate representation $\mathcal{F}_t^{tem''}$ like $\mathbb{S}$. Following the shared part, a fusion sub-net takes the concatenation of $\mathcal{F}_t^{spa''}$ and $\mathcal{F}_t^{tem''}$ as input, fuse them together, and infer the final spatiotemporal attention map $\mathcal{S}^t$
\begin{equation}
\mathcal{S}_t=f^{fuse}(\mathrm{cat}(\mathcal{F}_t^{spa''}, \mathcal{F}_t^{tem''})),
\end{equation}
where $f^{fuse}$ refers to the fusion sub-net. Note that the fusion sub-net has the same structure as the last five blocks except for its first block, which has doubled channels. The detail of $\mathbb{S}_{sp}$ is illustrated in Fig. \ref{fig:network} (b).
Different from $\mathbb{S}$, $\mathbb{S}_{sp}$ is trained by optimizing the spatiotemporal loss $\mathcal{L}_{sp}$, which is only related to the hard loss $\mathcal{L}_{hard}$
\begin{equation}
\mathcal{L}_{sp}=\mathcal{L}_{hard}(\mathbb{S}_{sp}(I_t,I_{t+1}),Y_t). \label{eq:l_sp}
\end{equation}

\subsection{Training Details}
All networks proposed in this paper are implemented with Tensorflow \cite{abadi2016tensorflow} on an NVIDIA GPU 1080Ti and a six-core CPU Intel 3.4GHz. In the knowledge distillation step, $\mathbb{S}$ is trained from scratch with a learning rate of $5\times 10^{-4}$ and a batch size of 96. Adam \cite{kingma2014adam} is employed to minimize the spatial loss $\mathcal{L}_{spa}$ and temporal loss $\mathcal{L}_{tmp}$. After training, the $\mathbb{S}$ performs well in extracting both spatial and temporal features from successive frame pair $(I_t, I_{t+1})$. In the spatiotemporal joint optimization step, the knowledge inherited from $\mathbb{S}$ is transferred into $\mathbb{S}_{sp}$, by optimizing the $\mathcal{L}_{sp}$, $\mathbb{S}_{sp}$ can perform well in extracting the spatiotemporal features.

\section{Experiments}
\label{sec:experiment}
We evaluate the proposed UVA-Net on a public dataset AVS1K~\cite{fu2018how}, which is an aerial video dataset for attention prediction.

\begin{table*}[t]
\footnotesize
\centering
\caption{Performance comparisons of 13 state-of-the-art models on AVS1K. The best and runner-up models of each column are marked with bold and underline, respectively. Except our model, the other deep models fine-tuned on AVS1K are marked with *.}
\label{tab:performance_AVS1K}
\resizebox{1.0\textwidth}!{
\begin{tabular}{l|l|ccccc|c|c|c|c|c}
\toprule
\multicolumn{2}{c|}{\multirow{3}{*}{Models}} &~\multirow{3}{*}{AUC}&~\multirow{3}{*}{sAUC}&~\multirow{3}{*}{NSS}&~\multirow{3}{*}{SIM}&~\multirow{3}{*}{CC} &~\multirow{3}{*}{\begin{tabular}{@{}c@{}}Input\\Resolution\end{tabular}}&~\multirow{3}{*}{\begin{tabular}{@{}c@{}}Parameters \\ (M)\end{tabular}} &~\multirow{3}{*}{\begin{tabular}{@{}c@{}}Memory \\ Footprint \\ (MB)\end{tabular}} &\multicolumn{2}{c}{Speed (fps)} \\ \cline{11-12}
\multicolumn{2}{c|}{} & & & & & & & & &~\begin{tabular}{@{}c@{}}GPU \\ (NVIDIA 1080Ti)\end{tabular}&~\begin{tabular}{@{}c@{}}CPU \\ (Intel 3.4GHz)\end{tabular} \tabularnewline
\midrule
\multirow{3}{*}{\textbf{H}}&~HFT            &~0.789 &~0.715 &~1.671 &~0.408 &~0.539 &~$128\times 128$&~--- &~--- &~--- &~7.6  \tabularnewline
 &~SP                                   &~0.781 &~0.706 &~1.602 &~0.422 &~0.520 &~$\max\{h,w\}=320$&~--- &~--- &~--- &~3.6     \tabularnewline
 &~PNSP                                &~0.787 &~0.634 &~1.140 &~0.321 &~0.370 &~$400\times 400$&~--- &~--- &~--- &~---   \tabularnewline
\midrule
\multirow{2}{*}{\textbf{NL}}&~SSD          &~0.737 &~0.692 &~1.564 &~0.404 &~0.503 &~$256\times 256$&~--- &~--- &~--- &~2.9   \tabularnewline
 &~LDS                              &~0.808 &~0.720 &~1.743 &~0.452 &~0.565 &~Original Size&~--- &~--- &~--- &~4.6    \tabularnewline
\midrule
\multirow{8}{*}{\textbf{DL}}&~eDN           &~0.855 &~0.732 &~1.262 &~0.289 &~0.417 &~---&~--- &~--- &~--- &~0.2    \tabularnewline
 &~iSEEL                      &~0.801 &~\underline{0.767} &~1.974 &~0.458 &~\underline{0.636} &~$224\times 224$&~--- &~--- &~--- &~---   \tabularnewline
 &~DVA$^*$                              &~\textbf{0.864} &~0.761 &~\textbf{2.044} &~\textbf{0.544} &~\textbf{0.658} &~$224\times 224$&~25.07 &~59.01 &~49 &~2.5    \tabularnewline
 &~SalNet$^*$                         &~0.797 &~\textbf{0.769} &~1.835 &~0.410 &~0.593 &~$224\times 224$&~25.81 &~43.22 &~28 &~1.5     \tabularnewline
 &~TSNet$^*$                         &~0.843 &~0.719 &~1.754 &~0.479 &~0.561 &~$224\times 224$&~25.81 &~43.22 &~--- &~---     \tabularnewline
 &~STS$^*$                             &~0.804 &~0.732 &~1.821 &~0.472 &~0.578 &~$320\times 320$&~41.25 &~86.94 &~17 &~0.9     \tabularnewline
 &~\textbf{UVA-DVA-32}                                          &~0.850 &~0.740 &~1.905 &~0.522 &~0.615  &~$32\times 32$&~\textbf{0.16} &~\textbf{0.68} &~\textbf{10,106} &~\textbf{404.3}   \tabularnewline
 &~\textbf{UVA-DVA-64}                                          &~\underline{0.856} &~0.748 &~\underline{1.981} &~\underline{0.540} &~0.635  &~$64\times 64$&~\underline{0.16} &~\underline{2.73} &~\underline{2,588} &~\underline{101.7}   \tabularnewline
\bottomrule
\end{tabular}}
\end{table*}
On the AVS1K, the UVA-Net is compared with 11 state-of-the-art models for video attention prediction, including:

1)~Three heuristic models (group denoted as [H]): HFT \cite{li2013visual}, SP \cite{li2014visual} and PNSP \cite{fang2014video}.

2)~Two non-deep learning models (group denoted as [NL]): SSD \cite{Li2015Finding} and LDS \cite{Fang2017Learning}.

3)~Six deep learning models (group denoted as [DL]): eDN \cite{Vig2014Large}, iSEEL \cite{tavakoli2017exploiting}, DVA \cite{wang2018deep}, SalNet \cite{Pan2016Shallow}, TSNet \cite{Pan2016Shallow} and STS \cite{bak2017spatio}.

With respect to the prior investigation of \cite{bylinskii2018different}, we adopt five evaluation metrics in the comparisons, including the traditional Area Under the ROC Curve (AUC), the shuffled AUC (sAUC), the Normalized Scanpath Saliency (NSS), the Similarity Metric (SIM)~\cite{hou2013visual}, and the Correlation Coefficient (CC)~\cite{borji2012boosting}.

\subsection{Comparison with the State-of-the-art Models}
\begin{figure*}[t]
\begin{center}
   \includegraphics[width=1.0\textwidth]{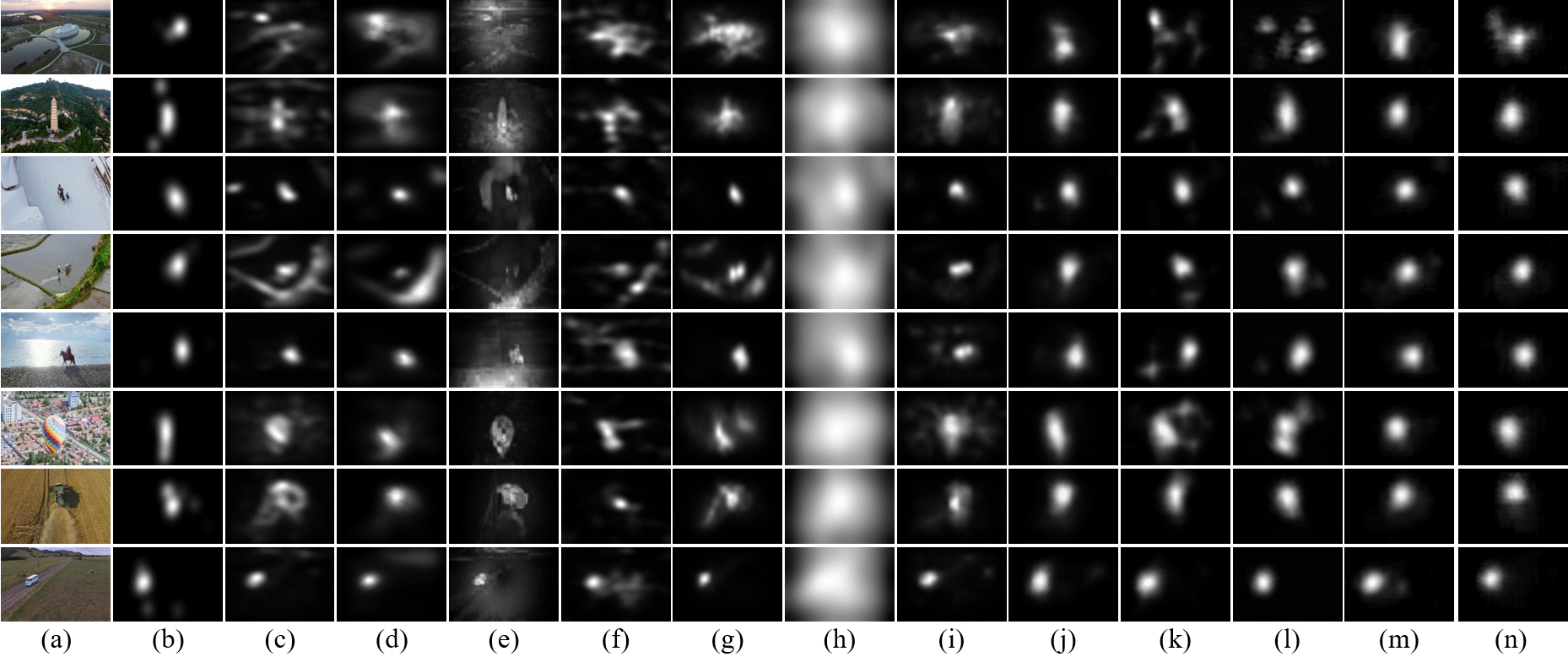}
\end{center}
   \caption{Representative frames of state-of-the-art models on AVS1K. (a) Video frame, (b) Ground truth, (c) HFT, (d) SP, (e) PNSP, (f) SSD, (g) LDS, (h) eDN, (i) iSEEL, (j) DVA, (k) SalNet, (l) STS, (m) UVA-DVA-32, (n) UVA-DVA-64.}
\label{fig:Result_AVS1K}
\end{figure*}

The performance of 13 state-of-the-art models on AVS1K is presented in Tab. \ref{tab:performance_AVS1K}. For sake of simplicity, we take UVA-Net in the resolution of $32\times 32$ and $64\times 64$, fix the spatial and temporal teacher models as DVA and TSNet \cite{bak2017spatio}, respectively. Some representative results of those models are illustrated in Fig. \ref{fig:Result_AVS1K}.

From Tab. \ref{tab:performance_AVS1K}, we observe that both the UVA-DVA-32 and UVA-DVA-64 are comparable to the 11 state-of-the-art models. In terms of AUC, NSS, and SIM, the UVA-DVA-64 ranks the second place and its CC ranks the third place while its sAUC ranks the fourth place. Such impressive performance can be attributed to the coupled knowledge distillation approach, which distills the spatial and temporal knowledge from well-trained complex teacher networks to a simple and compact student network, and finally transfers it into a spatiotemporal network. The knowledge distillation step makes it capable for the spatiotemporal network to separately extracting spatial and temporal cues from successive frame pair, while the spatiotemporal joint optimization step gives the spatiotemporal network the ability to fuse such spatial and temporal cues together to provide more powerful representations. Specially, we find that the UVA-DVA-64 is superior to traditional single branch networks (such as SalNet, with a $8.0\%$ performance gain), two-branch networks for video (such as STS, with a $8.8\%$ performance gain), but slightly inferior to multi-branch structure networks (such as DVA, with a $3.1\%$ performance drop). In addition, our UVA-Net has extremely low parameters (only $0.16$ M) and low memory footprint (UVA-DVA-32 takes $0.68$ MB and UVA-DVA-64 takes $2.73$ MB), resulting in high computational efficiency. In summary, the UVA-DVA-64 achieves very fast speed ($404.3$ FPS) and comparable performance to the state-of-the-art models. The UVA-DVA-32 achieves ultrafast speed ($10,106$ FPS) but with slight performance degradation.

We conduct a scalability experiment on DHF1K and present its performance in Tab. \ref{tab:generalization} to verify the proposed model's scalability to a ground-level dataset with normal viewpoints and target scales. Note that the DHF1K is the current largest ground-level video visual attention dataset and has made a significant leap in terms of scalability, diversity, and difficulty when compared with conventional ground-level datasets. On DHF1K, our model is compared with ten state-of-the-art models, including: PQFT~\cite{guo2010novel}, Seo \etal~\shortcite{seo2009static}, Rudoy \etal~\shortcite{rudoy2013learning}, Hou \etal~\shortcite{hou2009dynamic}, Fang \etal~\shortcite{fang2014video}, OBDL~\cite{hossein2015many}, AWS-D~\cite{leboran2016dynamic}, OM-CNN~\cite{Jiang_2018_ECCV}, Two-stream~\cite{bak2017spatio}. From this table, we find that our model is comparable with leading approaches, but runs two orders of magnitude faster.
\begin{table}[t]
\footnotesize
\centering
\resizebox{1.0\columnwidth}!{
\begin{threeparttable}
\caption{Performance comparison of 11 state-of-the-art dynamic models on DHF1K. The best and runner-up of each column are marked with bold and underline, respectively.} \label{tab:generalization}
\begin{tabular}{l@{}|ccccc|c}
 \toprule
 Model&~AUC&~sAUC&~NSS&~SIM&~CC&~\begin{tabular}{@{}c@{}}Speed\\(FPS)\end{tabular}\tabularnewline \midrule
  PQFT                                 &~0.699 &~0.562 &~0.749 &~0.139 &~0.137 &~--    \tabularnewline
  Seo \etal                            &~0.635 &~0.499 &~0.334 &~0.142 &~0.070 &~--    \tabularnewline
  Rudoy \etal                          &~0.769 &~0.501 &~1.498 &~0.214 &~0.285 &~--    \tabularnewline
  Hou \etal                            &~0.726 &~0.545 &~0.847 &~0.167 &~0.150 &~--    \tabularnewline
  Fang \etal                           &~0.819 &~0.537 &~1.539 &~0.198 &~0.273 &~--    \tabularnewline
  OBDL                                 &~0.638 &~0.500 &~0.495 &~0.171 &~0.117 &~--    \tabularnewline
  AWS-D                                &~0.703 &~0.513 &~0.940 &~0.157 &~0.174 &~--    \tabularnewline
  OM-CNN                               &~\underline{0.856} &~\underline{0.583} &~\underline{1.911} &~\underline{0.256} &~\underline{0.344} &~30    \tabularnewline
  Two-stream                           &~0.834 &~0.581 &~1.632 &~0.197 &~0.325 &~17    \tabularnewline
  ACL        &~\textbf{0.890} &~\textbf{0.601} &~\textbf{2.354} &~\textbf{0.315} &~\textbf{0.434} &~\underline{40}    \tabularnewline
  \textbf{OUR\tnote{*}}                    &~0.833 &~0.582 &~1.536 &~0.241 &~0.307 &~\textbf{2,588}\tabularnewline
\bottomrule
\end{tabular}
\begin{tablenotes}
    \footnotesize
    \item[*] Models test on the validation set of DHF1K.
\end{tablenotes}
\end{threeparttable}}
\end{table}

\subsection{Detailed Performance Analysis}
Beyond performance comparisons, we also conduct several experiments on AVS1K to verify the effectiveness of the proposed UVA-Net.

\noindent\textbf{Resolution reduction and supervision signal.} We conduct an experiment to assess the resolution reduction and supervision signal. Without loss of generality, we fix the temporal teacher signal as TSNet, and provide three candidate spatial teacher signals, DVA, SalNet and SSNet~\cite{bak2017spatio} in resolution $256\times 256$, $128\times 128$, $96\times 96$, $64\times 64$ and $32\times 32$. The performance of UVA-Net with different settings are presented in Tab. \ref{tab:distillation_AVS1K}.
\begin{table}[t]
\footnotesize
\centering
\caption{The performance comparisons of the UVA-Net with different settings on AVS1K dataset. T-S: spatial teacher model, Res: input resolution. The best model of each column in each spatial teacher signal are marked with bold.}
\label{tab:distillation_AVS1K}
\resizebox{1.0\columnwidth}!{
\begin{tabular}{c|c|ccccc}
\toprule
T-S&Res     &~AUC&~sAUC&~NSS&~SIM&~CC \tabularnewline
\midrule
\multirow{5}{*}{\begin{sideways}\textbf{UVA-DVA}\end{sideways}} &~256          &~0.786	&~0.680	 &~1.447	&~0.397	&~0.454    \tabularnewline
&~128          &~0.810	&~0.698	&~1.566	 &~0.438	&~0.498    \tabularnewline
&~96           &~0.827	&~0.726	&~1.765	 &~0.483	&~0.560    \tabularnewline
&~64           &~\textbf{0.856}	&~\textbf{0.748}	&~\textbf{1.981}	 &~\textbf{0.540}	&~\textbf{0.635}    \tabularnewline
&~32           &~0.850	&~0.740	&~1.905	 &~0.522	&~0.615    \tabularnewline
\midrule
\multirow{5}{*}{\begin{sideways}\textbf{UVA-SalNet}\end{sideways}} &~256          &~0.775	&~0.676	 &~1.423	&~0.394	 &~0.451   \tabularnewline
&~128          &~0.807	&~0.698	&~1.623	 &~0.451	&~0.511    \tabularnewline
&~96           &~0.843	&~0.717	&~1.746	 &~0.484	&~0.551    \tabularnewline
&~64           &~\textbf{0.859}	&~\textbf{0.751}	&~\textbf{1.955}	 &~\textbf{0.529}	&~\textbf{0.626}    \tabularnewline
&~32           &~0.852	&~0.739	&~1.892	 &~0.519	&~0.611    \tabularnewline
\midrule
\multirow{5}{*}{\begin{sideways}\textbf{UVA-SSNet}\end{sideways}}&~256          &~0.786	 &~0.687	&~1.485	 &~0.404	&~0.467    \tabularnewline
&~128          &~0.807	&~0.703	&~1.641	 &~0.455	&~0.517    \tabularnewline
&~96           &~0.834	&~0.721	&~1.788	 &~0.493	&~0.564    \tabularnewline
&~64           &~\textbf{0.852}	&~\textbf{0.744}	&~\textbf{1.971}	 &~\textbf{0.535}	&~\textbf{0.627}    \tabularnewline
&~32           &~0.845	&~0.736	&~1.894	 &~0.522	&~0.610    \tabularnewline
\bottomrule
\end{tabular}}
\end{table}
From this table, we find that models in low-resolution trends to have better performance. For example, with spatial supervision signal as DVA, the performance of UVA-Net in terms of all metrics in resolution $64\times 64$ ranks the first place. However, further resolution reduction will result in non-negligible performance degradation. We can infer that our UVA-Net can extract powerful spatiotemporal features from proper low-resolution videos, and it is still challenging for the UVA-Net in dealing with the details in high-resolution videos. In practical, compared with UVA-Net in $64\times 64$, the one with $256\times 256$ suffers from a $27.0\%$ performance drop.

Overall, the UVA-Net with DVA as teacher signal achieves the best performance, while SSNet and SalNet rank in the second and third places, respectively. For example, with resolution $64\times 64$, the UVA-Net supervised with DVA achieves NSS=$1.981$, while the SalNet and SSNet have only $1.955$ and $1.971$, respectively. This is consistent with the performance of teacher models in Tab. \ref{tab:performance_AVS1K}, which indicates the proposed approach can effectively transfer the knowledge inherited in teacher models.

\begin{figure}[t]
\begin{center}
   \includegraphics[width=1.0\linewidth]{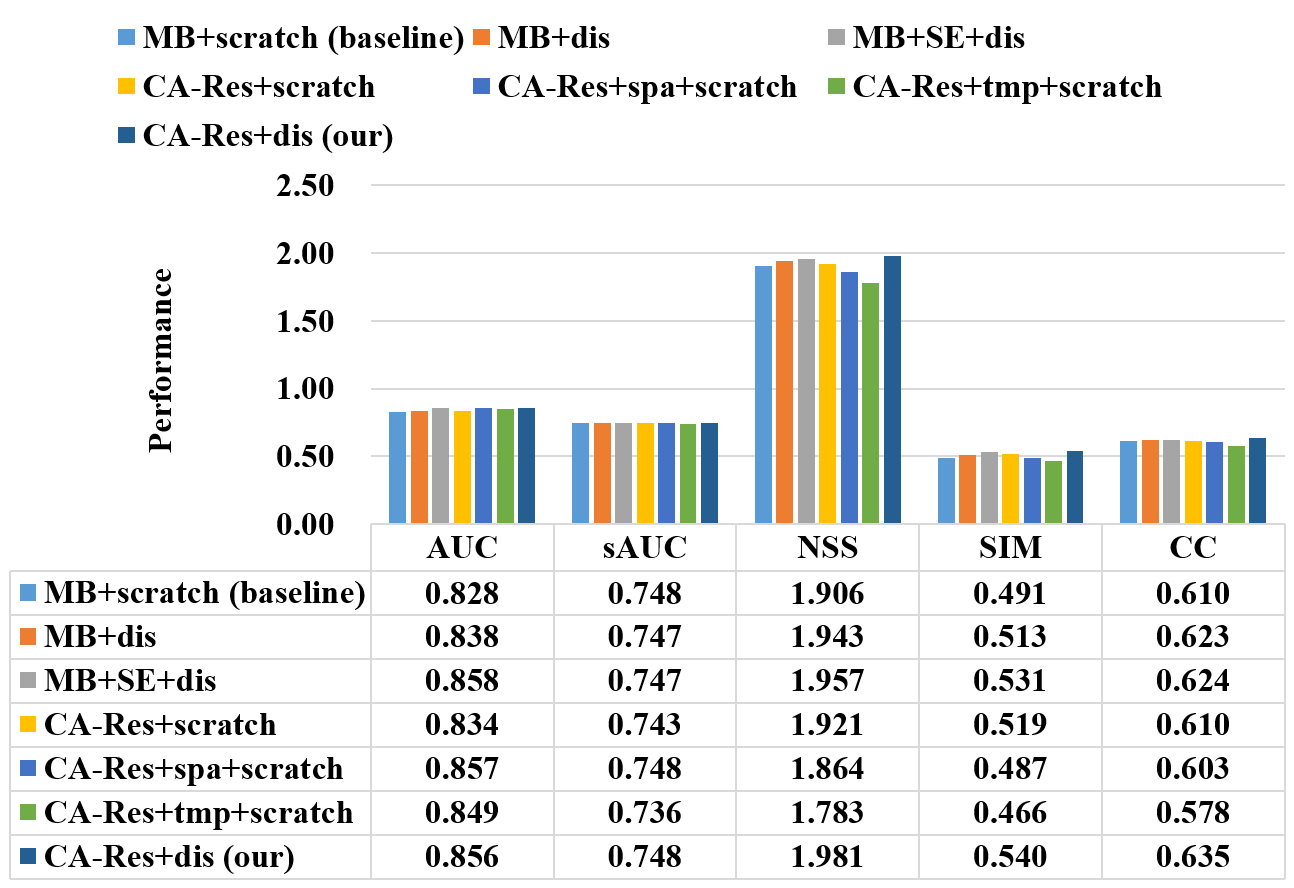}
\end{center}
   \caption{The performance comparisons of seven ablation models on AVS1K dataset.}
\label{fig:ablation}
\end{figure}

\noindent\textbf{Ablation study.} We use the AVS1K dataset and adopt the UVA-Net with MobileNet V2 blocks trained from scratch as the baseline, denoted as \textbf{MB+scratch}. We empirically show the effectiveness of our design choice via six experiments. 1) \textbf{MB+dis}. The UVA-Net with MobileNet V2 blocks with coupled knowledge distillation. 2) \textbf{MB+SE+dis}. The UVA-Net with MobileNet V2 blocks and SE block with coupled knowledge distillation. 3) \textbf{CA-Res+scratch}. The UVA-Net with CA-Res blocks trained from scratch. 4) \textbf{CA-Res+spa+scratch}. The student network spatial branch with CA-Res blocks trained from scratch. 5) \textbf{CA-Res+tmp+scratch}. The student network temporal branch with CA-Res blocks trained from scratch. 6) \textbf{CA-Res+dis}. The UVA-Net with CA-Res blocks with coupled knowledge distillation. The performances of all these ablation models are described in Fig. \ref{fig:ablation}.

From this figure, we find that \textbf{CA-res+scratch} achieves a performance gain to \textbf{MB+scratch}, \ie~NSS: $1.921\rightarrow 1.906$, and \textbf{CA-res+dis} is superior to \textbf{MB+dis}, \ie~NSS: $1.981\rightarrow 1.943$, which verify the effectiveness of the proposed channel-wise attention mechanism. Similarly, the effectiveness of our coupled knowledge distillation can be proved by the fact that \textbf{MB+dis} and \textbf{CA-Res+dis} are superior to \textbf{MB+scratch} and \textbf{CA-Res+scratch}, respectively.
\textbf{CA-Res+dis} has a $2.0\%$ performance gain to \textbf{MB+SE+dis}, \ie~NSS: $1.981\rightarrow 1.943$, which infers the superior of our channel-wise attention mechanism to traditional SE block. The reason behind this may be that the SE block adopts global average pooling and ignore global max pooling, making it challenging in maintaining the texture characteristic of feature maps. Additionally, without spatiotemporal joint optimization, \textbf{CA-Res+spa+dis} and \textbf{CA-Res+tem+dis} have a non-negligible performance degradation to \textbf{CA-Res+dis}, \ie~NSS: $1.864\rightarrow 1.981$ and $1.783\rightarrow 1.981$, respectively. This verifies that exploiting both spatial and temporal cues can greatly improve the representation power of networks than using each independently.

\begin{figure}[t]
\begin{center}
   \includegraphics[width=1.0\linewidth]{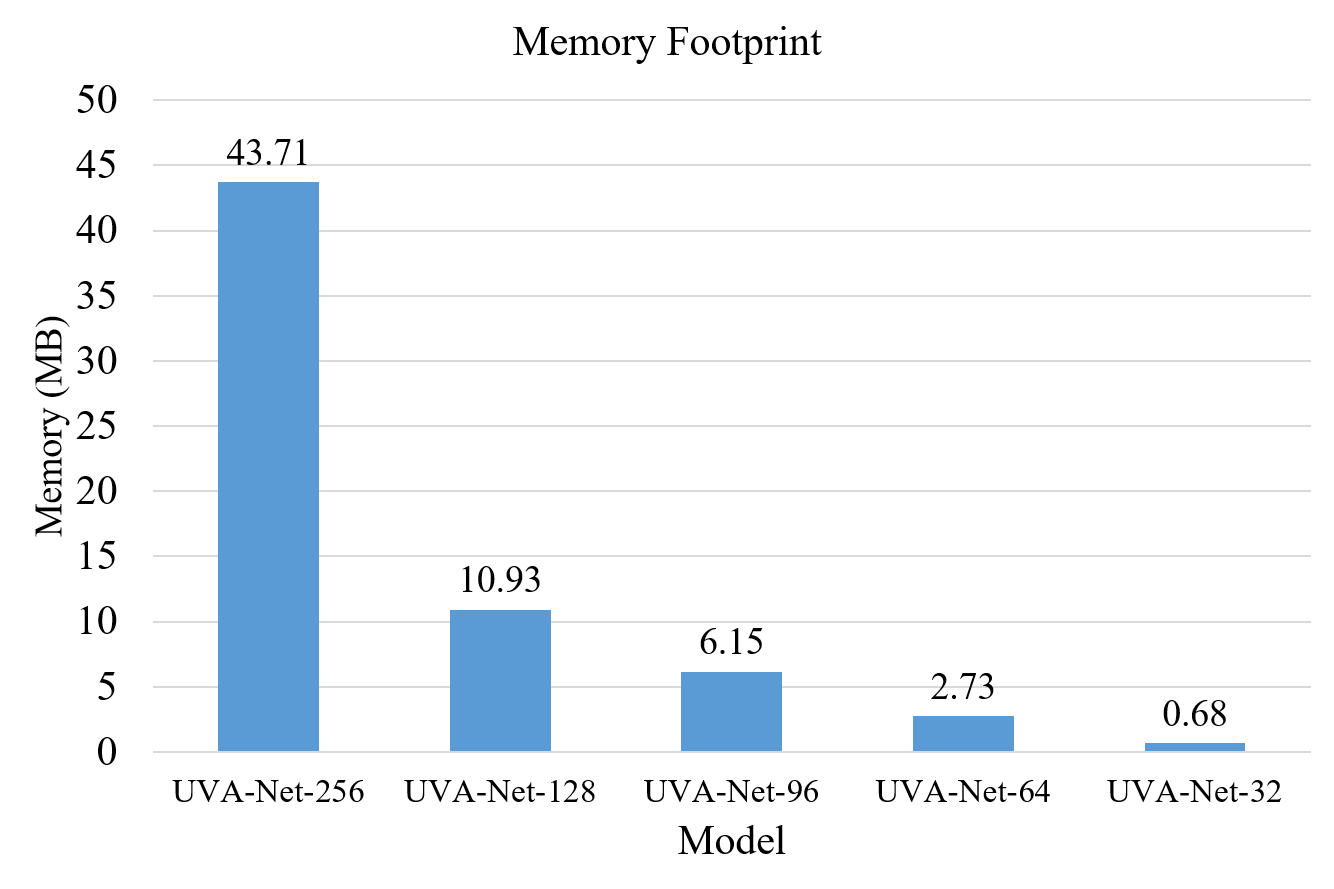}
\end{center}
   \caption{The inference memory footprint for various models. the memory footprint is proportional to the resolution of input.}
\label{fig:memory_footprint}
\end{figure}

\noindent\textbf{Speed analysis.} Our approach can greatly reduce the computational cost and memory space without remarkable loss of prediction accuracy. In particular, the DVA, SalNet, and STS contain $25.07$, $25.81$ and $41.25$ million parameters, while the proposed UVA-Net contains only $0.16$ million parameters. Namely, the UAV-Net achieves a substantial attenuation in the parameter amount. In addition, the memory footprints of UVA-Net in the different resolution are illustrated in Fig. \ref{fig:memory_footprint}.
We find that the memory footprint is proportional to the input resolution. For example, with a low resolution, such as $32\times 32$, the memory footprint of UVA-Net can be reduced to an extremely low value, $0.68$ MB. With such few parameters and low memory footprint, the process of attention inference can be greatly accelerated.

For comparison, we demonstrate the inference runtime of UVA-Net with different resolution on both high-end GPU (NVIDIA 1080Ti) and low-end CPU (Intel 3.4GHz) in Tab. \ref{tab:runtime}.
\begin{table}[t]
\footnotesize
\centering
\caption{Inference time and speed on GPU and CPU.}
\label{tab:runtime}
\resizebox{1.0\columnwidth}!{
\begin{tabular}{l|c|c}
\toprule
\multirow{2}{*}{Model} &~GPU~(NVIDIA 1080Ti) &~CPU~(Intel 3.4GHz) \tabularnewline
                       &~time/\#FPS &~time/\#FPS \tabularnewline
\midrule
UVA-Net-256 & 6.602~ms~/~151 & 134.230~ms~/~7.4 \tabularnewline
UVA-Net-128 & 1.586~ms~/~631 & 40.928~ms~/~24.4 \tabularnewline
UVA-Net-96 & 0.897~ms~/~1,115 & 22.495~ms~/~44.5 \tabularnewline
UVA-Net-64 & 0.386~ms~/~2,588 & 9.830~ms~/~101.7 \tabularnewline
UVA-Net-32 & 0.099~ms~/~10,106 & 2.474~ms~/~404.3 \tabularnewline
\bottomrule
\end{tabular}}
\end{table}
We observe that the inference runtime (with a NVIDIA 1080Ti) for attention prediction can be reduced to $6.602$ ms, $1.586$ ms, $0.897$ ms, $0.386$ ms and $0.099$ ms in resolution $256\times 256$, $128\times 128$, $96\times 96$, $64\times 64$ and $32\times 32$, respectively. Our model is a remarkable trade-off, which achieves ultrafast speed and impressive accuracy even with a CPU.

\section{Conclusion}
\label{sec:conclusion}
In this paper, we propose a simple yet powerful approach via coupled knowledge distillation for video attention prediction. In the knowledge distillation step, we distill the spatial and temporal knowledge inherited in the teacher networks to a student network. In the spatiotemporal joint optimization step, we transfer the knowledge learned by the student network to a spatiotemporal network and then fine-tune it. The proposed approach can greatly reduce the computational cost and memory space without remarkable loss of accuracy. The experimental results on a video dataset have validated the effectiveness of the proposed approach.

\section{ Acknowledgments}
This work was partially supported by grants from National Natural Science Foundation of China (61672072, 61922006 and 61532003), and the Beijing Nova Program (Z181100006218063).

{\small
\bibliographystyle{aaai}
\bibliography{kdbib}
}
\end{document}